\documentclass[11pt]{article}

\usepackage[margin=1in]{geometry}
\usepackage{amsmath,amssymb,amsfonts}
\usepackage{booktabs}
\usepackage{graphicx}
\usepackage{hyperref}
\usepackage{xcolor}

\newcommand{\R}{\mathbb{R}}
\newcommand{\polar}{\operatorname{polar}}

\title{How Much Orthogonalization Does Muon Need?}
\author{Hua Huang\thanks{\texttt{huah@nvidia.com}}\\
NVIDIA}

\begin{document}
\maketitle

\begin{abstract}
Muon optimizers improve neural-network training by replacing ill-conditioned
momentum updates with approximately semi-orthogonal updates.  This motivates
a practical question: how much orthogonalization does Muon actually require?  We
study this question using a relaxed cubic Newton--Schulz schedule derived
directly for Muon's low precision singular value band.  The resulting five-step
cubic construction uses ten dominant matrix multiplications, compared with
fifteen for five quintic Newton--Schulz iterations.  The cubic schedule is not
intended as a more accurate polar solver; instead, it is a principled low-cost
variant that lets us probe the relation between polar accuracy, spectral
shaping, and training quality.
Across synthetic diagnostics, NanoGPT ablations, and training experiments on
hybrid MoE/Mamba models, we find that training quality is not governed
monotonically by polar-decomposition accuracy: truncated Polar Express,
Muon-Jordan, cubic Newton--Schulz, and an explicit FP32 SVD polar factor can
reach nearly indistinguishable final loss on GPT-2 Small, and cubic5 matches
the Muon-Jordan quintic update within about $10^{-3}$ validation loss on 
hybrid MoE/Mamba models with one billion to four billion parameters.
These results support cubic5 as a practical low-cost Muon orthogonalization
variant, with empirical evidence of training-quality parity in the settings
tested.
\end{abstract}

\section{Introduction}

\subsection{The Muon Optimizer}

Muon is a newly proposed optimizer for matrix-valued hidden layer parameters in
neural networks \cite{jordan2024muon}.  It can be viewed as stochastic
gradient descent with momentum followed by an orthogonalization step.  Let
$W_t \in \R^{m \times n}$ be a weight matrix, $G_t$ its stochastic gradient, and
$M_t$ a momentum buffer.  A simplified Muon update is
\begin{align}
  M_t &= \beta M_{t-1} + (1-\beta) G_t, \\
  W_{t+1} &= W_t - \eta \, \polar(M_t).
\end{align}
If $M_t = U \Sigma V^\top$ is a singular value decomposition, then
\begin{equation}
  \polar(M_t) = U V^\top.
\end{equation}
Thus exact Muon would preserve the singular vectors of the momentum update but
replace all nonzero singular values by one.  Jordan et al. motivate this
operation empirically by observing that transformer momentum updates are often
nearly low-rank: a few directions dominate the update, while many ``rare''
directions have small singular values but may still matter for learning
\cite{jordan2024muon}.  Orthogonalization can therefore be understood as a
spectral reshaping operation that amplifies these suppressed directions.

This perspective is different from the standard numerical linear algebra
objective of computing a highly accurate polar factor.  In neural network
training, the update must remain useful after interaction with momentum,
learning rate schedules, normalization, weight decay, stochastic gradients, and
low precision arithmetic.  An exact SVD polar factor is the natural mathematical
reference point, but it need not optimize training loss.  The central
question of this work is therefore not merely whether a Newton--Schulz
iteration converges to the polar factor, but how accurately Muon must
orthogonalize its updates to preserve training quality.

The key observation behind these iterations is that odd matrix polynomials act
diagonally on singular values.  If $X = U \Sigma V^\top$ and
\begin{equation}
  X^+ = aX + b(XX^\top)X + c(XX^\top)^2 X,
\end{equation}
then
\begin{equation}
  X^+ = U \left(a\Sigma + b\Sigma^3 + c\Sigma^5\right) V^\top.
\end{equation}
The matrix problem can therefore be studied through scalar polynomial maps on
the singular values.  Classical polar decomposition
\cite{higham1986, higham1990, byers2008, nakatsukasa2010, higham2008, nakatsukasa2013, nakatsukasa2016}
algorithms and recently proposed hybrid polar decomposition method \cite{huang2024phd}
optimize these polynomials for high accuracy convergence to one.  
Muon, however, is used in a deep learning regime where low precision approximate
directions can be sufficient.  This creates room for polynomial schedules that
are less accurate as polar solvers but cheaper or differently biased as spectral
transforms.

\subsection{Polynomial Polar Decompositions for Muon}

The original Muon implementation uses five iterations of a fixed quintic
Newton--Schulz polynomial,
\begin{equation}
  p(x) = 3.4445x - 4.7750x^3 + 2.0315x^5,
\end{equation}
run in bfloat16 after Frobenius normalization \cite{jordan2024muon}.
We refer to this baseline as Muon-Jordan.  Its coefficients were chosen for the
relaxed Muon objective: after repeated composition, singular values are allowed
to lie in a band around one rather than converging to machine precision.
Liu et al. \cite{liu2025muon} further demonstrated that Muon can be scaled to 
larger Mixture-of-Experts (MoE) models containing 16 billion parameters.

Polar Express revisits this polynomial-design problem from the perspective of
matrix sign and polar decomposition methods \cite{amsel2026polarexpress}.  It uses
an adaptive sequence of quintic polynomials chosen by a minimax criterion.  This
improves the quality of the polar approximation and, when inserted into Muon,
can improve validation loss across learning rate sweeps.  Polar Express keeps
the same basic computational structure as standard quintic Newton--Schulz:
each iteration applies a degree-five odd polynomial using three dominant matrix
multiplications.

A complementary direction is to reduce the cost of the same polynomial
iteration by restructuring the computation.  Gram Newton--Schulz observes that
many Muon matrices are rectangular and that standard Newton--Schulz repeatedly
forms symmetric Gram matrices \cite{zhang2026gramnewton}.  By moving more work to the
smaller Gram matrix and using hardware-aware symmetric kernels, Gram
Newton--Schulz reduces the runtime of the orthogonalization step while
preserving the underlying polynomial iteration.  This is orthogonal to
coefficient design: one can improve the polynomial, the kernel, or both.

\subsection{Contributions}

This work uses a relaxed cubic construction to probe a broader question: how
much orthogonalization does Muon need, and what kind of spectral shaping matters
for training?  We make the following contributions.
\begin{itemize}
  \item We derive an adaptive cubic Newton--Schulz schedule whose coefficients
  are chosen from the current worst-case lower singular value bound and a
  relaxed target band $[0.7, 1.3]$.
  \item We identify a practical five-step schedule using the bfloat16 effective
  lower bound $l_0 = 7 \times 10^{-3}$.  This schedule uses 10 dominant matrix
  multiplications, compared with 15 for five quintic iterations.
  \item We compare cubic5, truncated Polar Express, truncated Muon-Jordan, and
  an explicit FP32 SVD polar factor inside NanoGPT training.  This separates
  polar-decomposition accuracy from optimizer quality: exact SVD does not
  improve over the strongest approximate Newton--Schulz updates in our GPT-2
  Small runs.
  \item We characterize a cost-quality frontier.  Cubic5 is close to Polar
  Express and SVD at the default GPT-2 Small setting, and it remains within
  approximately $10^{-3}$ validation loss of the Muon-Jordan quintic update on
  hybrid MoE/Mamba models with 1B to 4B parameters.  These results support
  cubic5 as a viable low-cost Muon orthogonalization variant in the evaluated
  regimes, not as a uniformly better update rule.
  \item We validate the scalar construction on synthetic matrices and show by
  microbenchmarking that reducing the polynomial degree lowers the cost of the
  orthogonalization subroutine.
\end{itemize}
Our claims are intentionally calibrated.  Cubic5 is not presented as a more
accurate polar decomposition algorithm than Polar Express, nor as a uniformly
better Muon update.  Rather, it is a practical low-cost Muon orthogonalization
variant whose empirical behavior also helps probe how polynomial-induced
spectral transforms affect training, beyond classical polar convergence alone.

\section{Relaxed Cubic Newton--Schulz for Muon Optimizer}

\subsection{Relaxed singular value Targets}

Classical polar decomposition aims to map every nonzero singular value to one.
Muon relaxes this requirement.  The original Muon coefficient search was guided
by the observation that training can tolerate singular values in a band such as
$[0.7, 1.3]$ \cite{jordan2024muon}.  This changes the polynomial-design problem:
the relaxed band is a useful design target, not a guarantee that training
quality is determined solely by whether every singular value lands in that
interval.

We work with the normalized iterate $X_0 = X/\|X\|_F$, so all singular values are
at most one.  In exact arithmetic, the smallest nonzero singular value may be
arbitrarily small.  Since Muon is used in a neural network training setting
with bfloat16 precision, a lower bound far below the relative precision is not
practically meaningful.  We therefore use
\begin{equation}
  l_0 = 7 \times 10^{-3} < \epsilon_{\text{bf16}} = 2^{-7} = 0.0078125
\end{equation}
as an effective lower bound.  This value produces a schedule that reaches the
relaxed lower target $0.7$ in five cubic iterations.
Muon-Jordan and Polar Express use a more conservative lower bound $l_0=10^{-3}$.
If we use the same lower bound, the schedule requires seven cubic iterations.

\subsection{Cubic Coefficient Derivation}

We derive the cubic coefficients using the same approach as Chen--Chow's scaled
cubic Newton--Schulz method \cite{chen2014}.
At iteration $t$, suppose all protected singular values lie in an interval
$[l_t,r_t]$.  We choose an odd cubic polynomial
\begin{equation}
  f_t(x) = a_t x + b_t x^3, \qquad b_t < 0,
\end{equation}
with peak value $u=1.3$.  Since $b_t<0$, the polynomial increases until its
critical point $k_t$ and then decreases.  We impose three conditions:
\begin{align}
  f_t'(k_t) &= 0, \\
  f_t(k_t) &= u, \\
  f_t(l_t) &= f_t(r_t).
\end{align}
The first two conditions set the shape once $k_t$ is known.  From
$f_t'(x)=a_t+3b_t x^2$,
\begin{equation}
  a_t = -3b_t k_t^2.
\end{equation}
Combining this with $f_t(k_t)=u$ gives
\begin{equation}
  b_t = -\frac{u}{2k_t^3}, \qquad
  a_t = \frac{3u}{2k_t}.
\end{equation}
The endpoint condition determines the peak location.  Substituting the above
form into $f_t(l_t)=f_t(r_t)$ yields
\begin{equation}
  3 l_t k_t^2 - l_t^3 = 3 r_t k_t^2 - r_t^3,
\end{equation}
and hence
\begin{equation}
  k_t^2 = \frac{r_t^2 + r_t l_t + l_t^2}{3}.
\end{equation}
Equivalently, if
\begin{equation}
  \alpha_t = \frac{1}{k_t}
  = \sqrt{\frac{3}{r_t^2+r_t l_t+l_t^2}},
\end{equation}
then
\begin{equation}
  f_t(x) = \frac{u}{2}\left(3\alpha_t x - \alpha_t^3 x^3\right).
\end{equation}

The endpoint equality is important.  Because the cubic has a single interior
maximum on the protected interval, the minimum over $[l_t,r_t]$ occurs at one
of the endpoints.  Enforcing $f_t(l_t)=f_t(r_t)$ balances the two endpoints and
maximizes the new worst-case lower bound under the chosen peak constraint.

\subsection{Adaptive Worst-Case Schedule}

The coefficients are recomputed at every step from the current lower bound.  We
initialize
\begin{equation}
  l_0 = 7 \times 10^{-3}, \qquad r_0 = 1,
\end{equation}
and after the first step protect the relaxed upper range by setting
\begin{equation}
  r_t = u = 1.3, \qquad t \ge 1.
\end{equation}
The worst-case lower bound evolves as
\begin{equation}
  l_{t+1} = f_t(l_t).
\end{equation}
For $l_0=7\times10^{-3}$, this construction reaches $l_t \ge 0.7$ after five
steps.  The resulting cubic5 coefficients are
\begin{center}
\begin{tabular}{rrrr}
\toprule
$t$ & $a_t$ & $b_t$ & $l_{t+1}$ \\
\midrule
0 & 3.3656576 & -3.3420992 & 0.0235585 \\
1 & 2.5744352 & -1.4957376 & 0.0606302 \\
2 & 2.5368962 & -1.4312570 & 0.1534934 \\
3 & 2.4418906 & -1.2764040 & 0.3701983 \\
4 & 2.2230472 & -0.9630650 & 0.7741077 \\
\bottomrule
\end{tabular}
\end{center}

The corresponding matrix iteration is
\begin{equation}
  X_{t+1} = a_t X_t + b_t (X_tX_t^\top)X_t,
\end{equation}
with the usual transpose trick for tall matrices so that the smaller Gram
matrix is formed.  This preserves singular vectors and applies $f_t$ to each
singular value.

\begin{figure}[htb]
  \centering
  \includegraphics[width=0.95\linewidth]{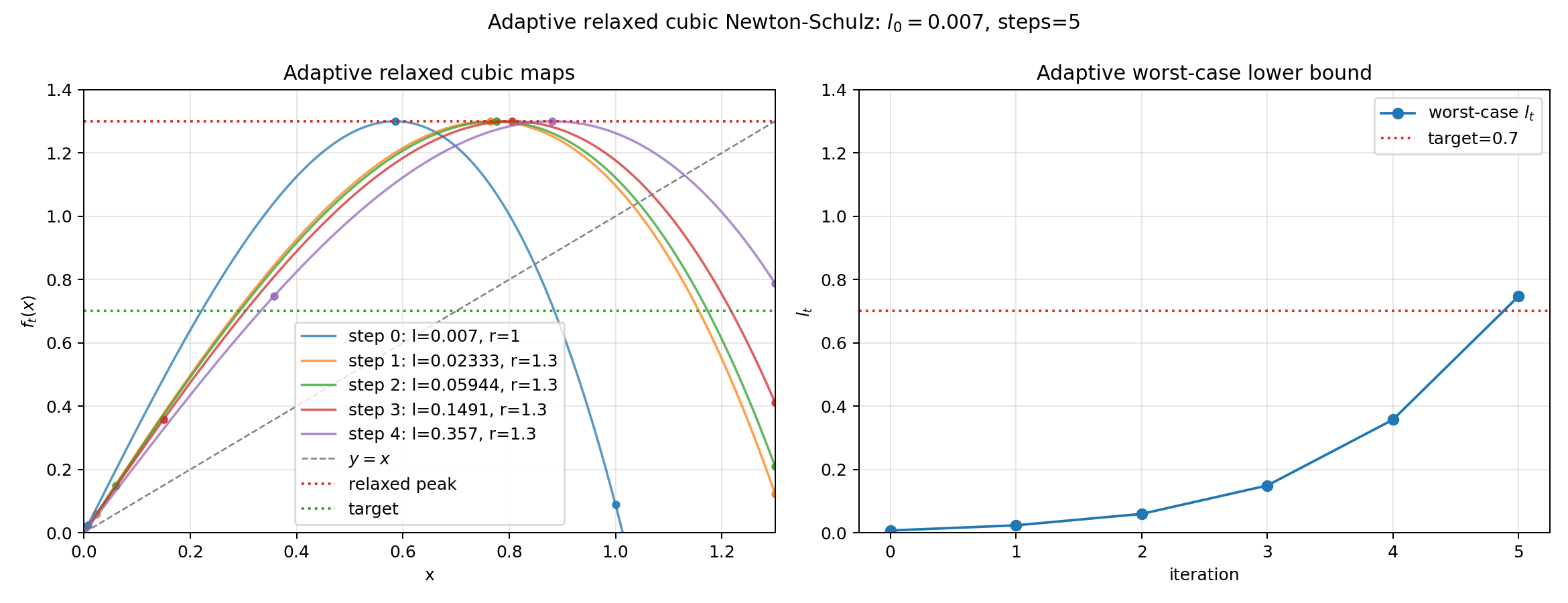}
  \caption{Adaptive relaxed cubic Newton--Schulz schedule for the effective
  bfloat16 lower bound $l_0=7\times 10^{-3}$.  The coefficients are recomputed
  from the current worst-case lower bound, and the scalar lower-bound trajectory
  reaches the relaxed target band in five cubic iterations.}
  \label{fig:relaxed-cubic-schedule}
\end{figure}

\subsection{FLOP and Matrix-Multiplication Count}

We count only dominant matrix multiplications and ignore scalar multiplications,
matrix additions, and normalization.  A cubic step
\begin{equation}
  X_{t+1} = a_t X_t + b_t (X_tX_t^\top)X_t
\end{equation}
requires two dominant multiplications: one to form the Gram matrix and one to
multiply it back into $X_t$.  Therefore five cubic steps require ten dominant
matrix multiplications.

A quintic Newton--Schulz step of the form
\begin{equation}
  X_{t+1} =
  a_t X_t + b_t (X_tX_t^\top)X_t + c_t (X_tX_t^\top)^2X_t
\end{equation}
requires three dominant multiplications: one for the Gram matrix, one for the
Gram-square or polynomial-in-Gram term, and one to multiply the result by
$X_t$.  Five quintic steps, including Muon-Jordan and Polar Express, therefore
require fifteen dominant multiplications.

For an $m\times n$ matrix with $m\ge n$, each cubic step has leading cost
approximately
\begin{equation}
  2mn^2 + 2mn^2 = 4mn^2
\end{equation}
floating point operations when the smaller $n\times n$ Gram matrix is used.  A
quintic step adds an $n\times n$ multiplication, yielding roughly
\begin{equation}
  4mn^2 + 2n^3.
\end{equation}
For highly rectangular matrices, the rectangular multiplications dominate and
the cubic schedule saves roughly one third of the dominant multiplications.  For
near-square matrices, the $n^3$ Gram-square term is also substantial, so the
benefit remains meaningful.  Actual wall-clock speedups depend on batching,
kernel fusion, symmetric-kernel use, matrix aspect ratio, and whether the
orthogonalization subroutine is a large fraction of the optimizer step
\cite{zhang2026gramnewton}.

\section{Numerical Results}

The experiments are organized around two questions.  First, does the relaxed
cubic derivation produce the expected spectral transform and subroutine cost
reduction?  Second, how closely does training quality track polar accuracy once
the orthogonalization routine is inserted into Muon?  This second question is
central to the optimizer setting: an update can be numerically less accurate as
a polar factor while still being similarly useful for training.

\subsection{Singular Value Diagnostics and Orthogonalization Microbenchmark}

The first experiment checks whether the scalar construction transfers to matrix
polynomials in low precision.  We construct an FP32 square matrix
$M=U\operatorname{diag}(s)V^\top$ with controlled singular values
$s=\operatorname{logspace}(10^{-5},1)$, cast it to bfloat16, run the matrix
polynomial, and compute output singular values in FP32.  The diagnostic compares
cubic5 in bfloat16, cubic5 in FP32, Polar Express in bfloat16, and Muon-Jordan
in bfloat16.

\begin{figure}[htb]
  \centering
  \includegraphics[width=\linewidth]{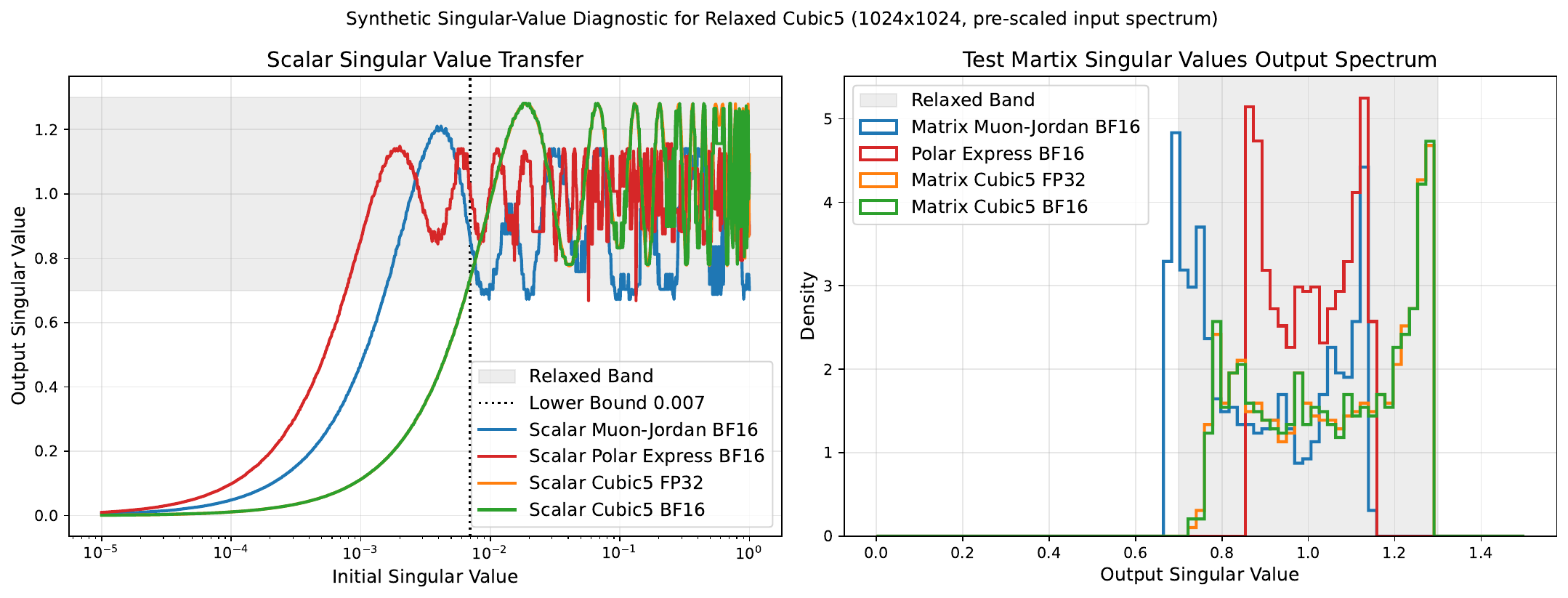}
  \caption{Synthetic singular value diagnostic.  Left: scalar polynomial
  transfer on a wide range of input singular values.  Right: output singular
  value spectrum for matrices whose input singular values lie in the protected
  interval $[7\times 10^{-3},1]$.}
  \label{fig:synthetic-sv-diagnostic}
\end{figure}

The result supports the intended interpretation of the bfloat16 lower bound.
The method does not claim that singular values below $7\times10^{-3}$ are
guaranteed to enter the relaxed band.  Instead, singular values at or above this
effective lower bound are mapped stably into the target region, while much
smaller singular values do not produce numerical blow-up or contaminate the
output spectrum.  The FP32 and bfloat16 cubic curves are close on the protected
range, which indicates that the scalar derivation remains predictive for the
matrix implementation.

We also benchmark the orthogonalization subroutine on representative matrix
shapes on an NVIDIA H200 GPU.  The microbenchmark shows that the cubic5 routine
reduces the cost of the orthogonalization step relative to Polar Express.  
This is consistent with the dominant multiplication count: both methods use the
same normalization and general Newton--Schulz structure, but cubic5 avoids the
Gram-square multiplication in each iteration.  We treat this as a subroutine-level
result, not as a claim of proportional end-to-end training speedup.

\begin{table}[htb]
  \centering
  \begin{tabular}{llllr}
    \toprule
    Model & Layer or weight matrix & Shape & Matrix size & Speedup \\
    \midrule
    Kimi K2 & Expert MLP up/gate/down & rectangular & $2048 \times 7168$ & $1.28\times$ \\
    Kimi K2 & Dense MLP up/gate/down & rectangular & $7168 \times 18432$ & $1.23\times$ \\
    Llama 3 70B & GQA key/value projection & rectangular & $1024 \times 8192$ & $1.30\times$ \\
    Llama 3 70B & Query/output projection & square & $8192 \times 8192$ & $1.52\times$ \\
    Llama 3 70B & MLP up/gate/down & rectangular & $8192 \times 28672$ & $1.16\times$ \\
    \bottomrule
  \end{tabular}
  \caption{Orthogonalization-subroutine microbenchmark.  Matrix shapes follow
  the Kimi K2 and Llama 3 70B examples used in the Gram Newton--Schulz
  discussion.  Speedup is the median runtime ratio of Polar Express over
  relaxed cubic5 for the same synthetic gradient matrix.}
  \label{tab:ortho-microbenchmark}
\end{table}

\subsection{Orthogonalization Accuracy vs. Training Quality}

\begin{figure}[htb]
  \centering
  \includegraphics[width=0.85\linewidth]{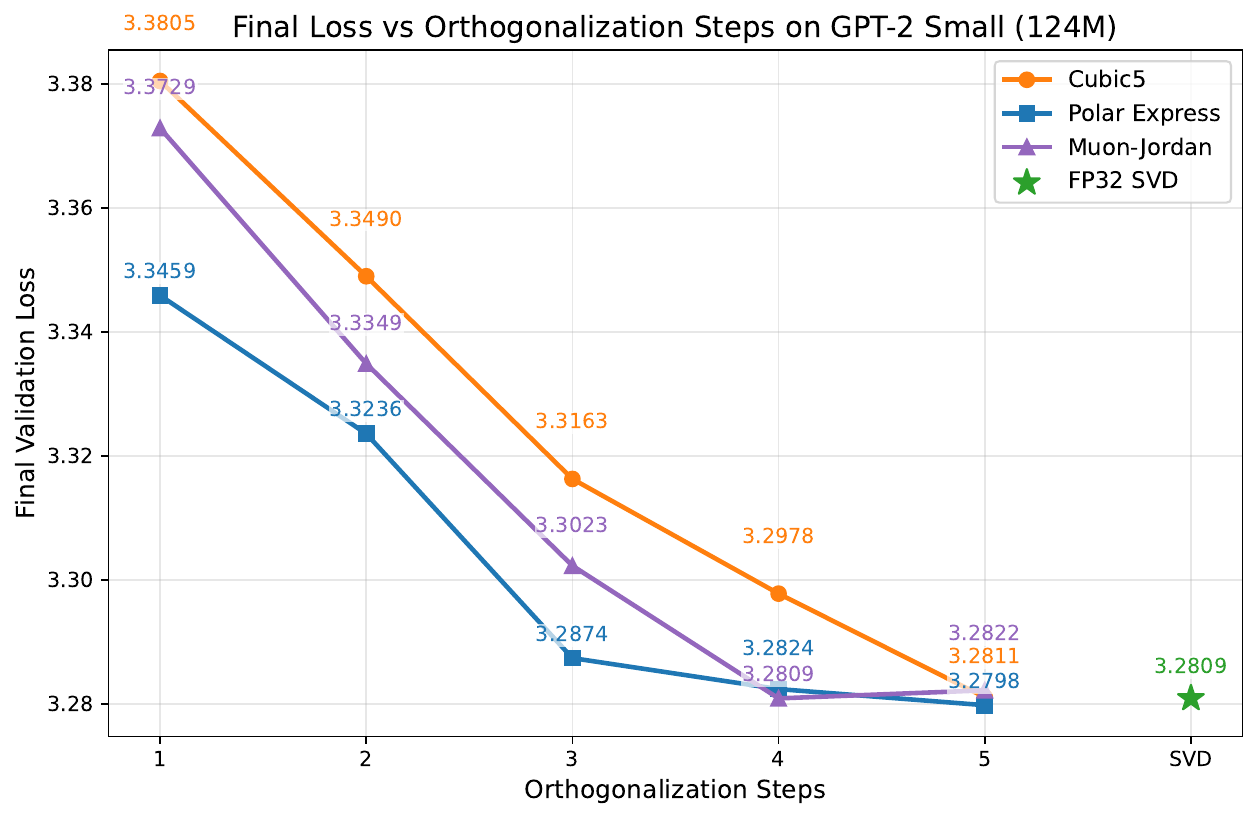}
  \caption{Final validation loss versus orthogonalization steps on GPT-2
  Small.  Cubic5, Polar Express, and Muon-Jordan all approach the same quality
  region as more steps are added, while the explicit FP32 SVD polar factor does
  not improve over the strongest approximate Newton--Schulz updates.}
  \label{fig:orthogonalization-frontier}
\end{figure}

The central training ablation compares four ways of producing a Muon update on
GPT-2 Small: the relaxed cubic schedule truncated to one through five steps,
Polar Express truncated to one through five steps, Muon-Jordan truncated to one
through five steps, and an explicit FP32 SVD polar factor.  The SVD baseline
computes the polar factor most directly, but it is much too expensive to be a
practical Muon routine.  Its purpose here is conceptual: it distinguishes exact
polar orthogonalization from the update that achieves the best validation loss.
Figure~\ref{fig:orthogonalization-frontier} shows the result for random seed
$0$.  We repeated the same ablation with seeds $42$ and $33550336$ and observed
the same qualitative pattern.

This comparison is informative.  Cubic5 improves monotonically from final loss
$3.3805$ with one step to $3.2811$ with five steps.  Polar Express also improves
with additional steps, reaching $3.2798$ at five steps.  Muon-Jordan improves
rapidly through four steps, from $3.3729$ to $3.2809$, and then slightly
increases to $3.2822$ at five steps.  The gap among the strongest variants is
very small: five-step cubic5 reaches $3.2811$, four-step Polar Express reaches
$3.2824$, five-step Polar Express reaches $3.2798$, four-step Muon-Jordan
reaches $3.2809$, and the explicit FP32 SVD polar factor also reaches
$3.2809$.  These runs differ by less than $0.003$ final validation loss, while
their numerical interpretation and matrix-multiplication costs are quite
different.

This ablation suggests that optimizer quality is not determined solely by polar
decomposition accuracy.  If polar accuracy were the governing factor, the SVD
baseline would be expected to dominate the approximate polynomial methods.
Instead, training quality appears to depend on the effective spectral transform
applied to the momentum update.  Polar Express shapes the spectrum more
effectively than cubic5 at very small step counts, but by five cubic steps the
cheaper cubic5 schedule reaches the same quality region.  The non-monotone
five-step Muon-Jordan result further indicates that adding another approximate
orthogonalization step need not monotonically improve training loss.  We
therefore view cubic5 as a principled point on a cost-quality frontier rather
than as a method that dominates truncated quintic or exact SVD baselines.

\subsection{Nemotron-3-Nano Hybrid MoE/Mamba Scaling Experiments}

The NanoGPT ablation in Figure~\ref{fig:orthogonalization-frontier} isolates the
effect of the orthogonalization routine in a controlled small-model setting.  We
next test whether the same relaxed cubic update remains viable when training
larger Nemotron-3-Nano models \cite{nvidia2025nemotron-h} with Megatron-LM.  
We use three hybrid MoE/Mamba configurations: 1B-A315M, 2B-A500M, and 4B-A770M
(with 1B/2B/4B total parameters and 315M/500M/770M active parameters, respectively).
Their core architectural hyperparameters are listed in 
Table~\ref{tab:nemotron-nano-architecture}.  All three models use grouped
query attention with 2 query groups, 128 experts, top-6 routing, a sigmoid
router score, and sequence-level auxiliary load-balancing loss with coefficient
$10^{-4}$.
We use 88B/140B/215B token subsets of the full Nemotron-3 dataset
\cite{nvidia2025nemotron-h} for the training.  Training uses sequence length
8192, bfloat16 arithmetic, global batch size 768, Muon optimizer, and a
warmup-stable-decay learning-rate schedule \cite{hagele2024} with a
minus-square-root decay style for the final WSD decay.  The token budgets and
learning-rate ranges are summarized in Table~\ref{tab:nemotron-nano-training}.

\begin{table}[htb]
  \centering
  \begin{tabular}{lrrrrrr}
    \toprule
    \multicolumn{1}{c}{Model} & \multicolumn{1}{c}{Layers} & \multicolumn{1}{c}{Hidden} & \multicolumn{1}{c}{Attn} & \multicolumn{1}{c}{Mamba} & \multicolumn{1}{c}{FFN} & \multicolumn{1}{c}{Shared} \\
    & & \multicolumn{1}{c}{size} & \multicolumn{1}{c}{heads} & \multicolumn{1}{c}{heads} & \multicolumn{1}{c}{size} & \multicolumn{1}{c}{expert size} \\
    \midrule
    1B-A315M & 19 & 768  & 6  & 24 & 512 & 960  \\
    2B-A500M & 24 & 1024 & 8  & 32 & 640 & 1280 \\
    4B-A770M & 30 & 1280 & 12 & 40 & 768 & 1536 \\
    \bottomrule
  \end{tabular}
  \caption{Core architectural hyperparameters for the Nemotron-3-Nano hybrid
  MoE/Mamba scaling experiments.  All three models use 128 experts and top-6 routing.
  The FFN column denotes the per-expert hidden size, and the shared-expert
  column denotes the shared expert intermediate size.}
  \label{tab:nemotron-nano-architecture}
\end{table}

\begin{table}[htb]
  \centering
  \begin{tabular}{lrrr}
    \toprule
    Model & Trained tokens & Max LR & Min LR \\
    \midrule
    1B-A315M & 88B  & $1.99{\times}10^{-3}$ & $1.99{\times}10^{-5}$ \\
    2B-A500M & 140B & $1.28{\times}10^{-3}$ & $1.28{\times}10^{-5}$ \\
    4B-A770M & 215B & $8.50{\times}10^{-4}$ & $8.50{\times}10^{-6}$ \\
    \bottomrule
  \end{tabular}
  \caption{Training token budgets and learning-rate ranges for the
  Nemotron-3-Nano model scaling experiments.}
  \label{tab:nemotron-nano-training}
\end{table}

\begin{table}[htb]
  \centering
  \begin{tabular}{lrrrrrr}
    \toprule
    \multicolumn{1}{c}{Model} & \multicolumn{3}{c}{Validation loss} & \multicolumn{3}{c}{PPL loss} \\
    \cmidrule(lr){2-4}\cmidrule(lr){5-7}
    & \multicolumn{1}{c}{Muon-Jordan} & \multicolumn{1}{c}{Cubic5} & \multicolumn{1}{c}{$\Delta$}
    & \multicolumn{1}{c}{Muon-Jordan} & \multicolumn{1}{c}{Cubic5} & \multicolumn{1}{c}{$\Delta$} \\
    \midrule
    1B-A315M & 1.609511 & 1.608741 & -0.000770 & 5.000363 & 4.996515 & -0.003848 \\
    2B-A500M & 1.496790 & 1.496687 & -0.000103 & 4.467325 & 4.466865 & -0.000460 \\
    4B-A770M & 1.412247 & 1.413246 & +0.000999 & 4.105170 & 4.109273 & +0.004103 \\
    \bottomrule
  \end{tabular}
  \caption{Final validation metrics for Nemotron-3-Nano models
  trained with Megatron-LM.  The Muon-Jordan baseline uses the fixed quintic
  five-step Newton--Schulz polynomial, while cubic5 uses the relaxed cubic
  coefficients.  $\Delta$ is cubic5 minus Muon-Jordan; negative values favor
  cubic5.}
  \label{tab:megatron-scaling}
\end{table}

\begin{figure}[htb]
  \centering
  \includegraphics[width=0.9\linewidth]{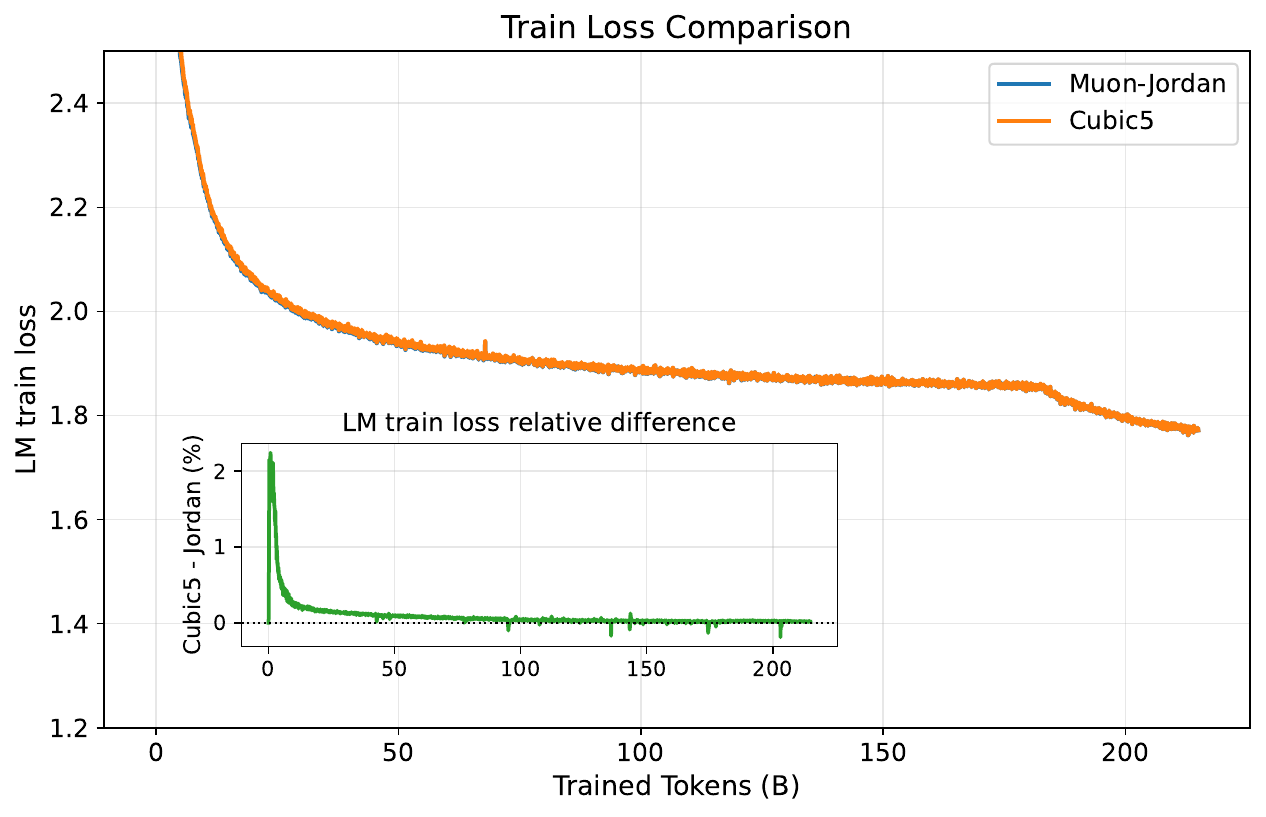}
  \caption{Training loss comparison for the Nemotron-3-Nano-4B-A770M model.  The
  inset shows the relative difference between cubic5 and Muon-Jordan using the
  same trained-token axis.}
  \label{fig:nemotron-4b-train-loss}
\end{figure}

The final validation losses are nearly indistinguishable at all three scales.
For the 1B and 2B models, cubic5 is slightly lower than the Muon-Jordan quintic 
update on both validation loss and PPL loss.  For the 4B model, cubic5 is slightly
higher, by about $10^{-3}$ validation loss and $0.004$ PPL loss.  The
corresponding test-set metrics show the same qualitative pattern: cubic5 is lower
by $0.000719$ and $0.000096$ test loss on the 1B and 2B models respectively, 
and higher by $0.000936$ test loss on the 4B model.

The 4B training curve in Figure~\ref{fig:nemotron-4b-train-loss} provides a
trajectory-level view of the largest scaling experiment.  The relative
difference is largest (about 2\%) at the beginning of training, but it decays
rapidly and then fluctuates around zero for the remaining token budget.  
This pattern is consistent with a transient difference in the early optimization
dynamics rather than a persistent shift in the training trajectory.  After the
initial phase, cubic5 and Muon-Jordan enter the same loss regime.  The
late-stage behavior is therefore consistent with the final validation metrics:
cubic5 does not improve over Muon-Jordan on the 4B model, but it preserves
essentially the same training trajectory while using a lower-degree
orthogonalization polynomial. The 1B and 2B model training curves are 
very similar to the 4B model and we omit them for brevity.

Taken together, these experiments provide the main training-scale evidence in
this draft.  Cubic5 does not uniformly improve over Muon-Jordan, but it reaches
comparable validation metrics in these scaling experiments while using a
lower-degree orthogonalization polynomial.  This supports cubic5 as a viable
low-cost Muon orthogonalization variant in regimes where the orthogonalization
subroutine is a relevant part of optimizer cost.  It also reinforces the smaller
ablation's message that, once the spectral transform is in a suitable regime,
training quality can be relatively insensitive to the exact polynomial used for
approximate orthogonalization.

\section{Discussion and Limitations}

The main conclusion is that Muon's orthogonalization requirement is an optimizer
requirement, not simply a polar-decomposition accuracy requirement.  The relaxed
cubic schedule was derived from a worst-case singular value target, but the
training experiments show that this target is neither a necessary nor a
sufficient description of optimizer quality.  Four-step Polar Express and
four-step Muon-Jordan can work well even though their scalar behavior differs
from both exact SVD and the full five-step schedules, and an explicit FP32 SVD
polar factor does not improve over the strongest approximate Newton--Schulz
updates on GPT-2 Small.  This suggests
that Muon benefits from a broader spectral reshaping effect: ill-conditioned
momentum updates are made more isotropic, but the exact singular value
map that best serves training depends on the training phase, learning rate, and
real update spectra.

The evidence therefore positions cubic5 as a principled low-cost point on the
Muon orthogonalization frontier.  Its advantage is computational: it removes one
dominant matrix multiplication from each Newton--Schulz step by using a cubic
instead of a quintic polynomial, giving a five-step routine with ten dominant
matrix multiplications.  The Nemotron-3-Nano experiments suggest that cubic5 can
match the Muon-Jordan quintic update closely on larger hybrid MoE/Mamba models,
but the differences are small enough that they should be interpreted as
equivalence-level evidence rather than as a quality advantage.  This motivates
further evaluation of cubic5 as a low-cost Muon orthogonalization variant,
without claiming universal dominance over Muon-Jordan or Polar Express.

The SVD comparison also changes how future Muon polynomial work should be
evaluated.  Exact polar decomposition is the natural mathematical endpoint, but
it is not a reliable proxy for training quality.  More informative diagnostics should
measure how a candidate update compares to strong Muon baselines on the spectra
that actually occur during training: update cosine similarity, Frobenius
relative error to Polar Express or SVD, amplification of bulk versus tail
singular directions, and how these quantities evolve during the early phase of
training.  Such measurements would connect scalar polynomial design to optimizer
dynamics more directly than asking only whether all singular values enter a
fixed relaxed band.

Finally, the current evidence separates subroutine speed from end-to-end
training speed.  Reducing the Newton--Schulz multiplication count lowers the
cost of the orthogonalization routine, but the full optimizer step includes
communication, parameter updates, data movement, and non-Muon parameters.  The
Nemotron-3-Nano experiments in Table~\ref{tab:megatron-scaling} primarily establish
training-quality parity; they should not yet be read as an end-to-end throughput
study.  End-to-end speedups will depend on model architecture, matrix aspect
ratios, hardware kernels, and parallelism.  Future work should combine relaxed
polynomial schedules with hardware-aware implementations such as Gram
Newton--Schulz and measure full training wall-clock time at larger scales.

\section*{Acknowledgements}

The author sincerely thanks Wenkai Shao, Mikail Khona, Hao Wu, and Jin Zhou for
their valuable discussions and assistance with the experiments.

\bibliographystyle{plain}
\bibliography{ref}

@misc{jordan2024muon,
    author = {Keller Jordan and Yuchen Jin and Vlado Boza and Jiacheng You and Franz Cesista and Laker Newhouse and Jeremy Bernstein},
    title = {Muon: An optimizer for hidden layers in neural networks},
    year = {2024},
    url = {https://kellerjordan.github.io/posts/muon/}
}

@misc{liu2025muon,
    title = {Muon is Scalable for {LLM} Training}, 
    author = {Jingyuan Liu and Jianlin Su and Xingcheng Yao and Zhejun Jiang and Guokun Lai and Yulun Du and Yidao Qin and Weixin Xu and Enzhe Lu and Junjie Yan and Yanru Chen and Huabin Zheng and Yibo Liu and Shaowei Liu and Bohong Yin and Weiran He and Han Zhu and Yuzhi Wang and Jianzhou Wang and Mengnan Dong and Zheng Zhang and Yongsheng Kang and Hao Zhang and Xinran Xu and Yutao Zhang and Yuxin Wu and Xinyu Zhou and Zhilin Yang},
    year = {2025},
    eprint = {2502.16982},
    archivePrefix = {arXiv},
    url = {https://arxiv.org/abs/2502.16982}, 
}

@inproceedings{amsel2026polarexpress,
    title = {The Polar Express: Optimal Matrix Sign Methods and their Application to the Muon Algorithm},
    author = {Noah Amsel and David Persson and Christopher Musco and Robert M. Gower},
    booktitle = {The Fourteenth International Conference on Learning Representations},
    year = {2026},
    url = {https://openreview.net/forum?id=yRtgZ1K8hO}
}

@misc{zhang2026gramnewton,
    title = {{G}ram {N}ewton-{S}chulz},
    author = {Jack Zhang and Noah Amsel and Berlin Chen and Tri Dao},
    year = {2026},
    url = {https://dao-ailab.github.io/blog/2026/gram-newton-schulz/}
}

@article{nakatsukasa2010,
    author = {Nakatsukasa, Yuji and Bai, Zhaojun and Gygi, Fran\c{c}ois},
    title = {Optimizing {Halley}'s Iteration for Computing the Matrix Polar Decomposition},
    year = {2010},
    publisher = {SIAM},
    address = {USA},
    volume = {31},
    number = {5},
    doi = {10.1137/090774999},
    journal = {SIAM Journal on Matrix Analysis and Applications},
    month = {9},
    pages = {2700--2720},
    numpages = {21},
}

@article{byers2008,
    title = {A New Scaling for {Newton}'s Iteration for the Polar Decomposition and its Backward Stability},
    volume = {30},
    doi = {10.1137/070699895},
    number = {2},
    urldate = {2022-01-11},
    journal = {SIAM Journal on Matrix Analysis and Applications},
    author = {Byers, Ralph and Xu, Hongguo},
    month = {1},
    year = {2008},
    pages = {822--843},
}

@online{chen2014,
    title = {A stable scaling of {Newton}-{Schulz} for improving the sign function computation of a {Hermitian} matrix},
    author = {Chen, Jie and Chow, Edmond},
    year = {2014},
    pages = {23},
    note = "Preprint ANL/MCS-P5059-0114, Argonne National Laboratory",
}

@book{higham2008,
    author = {Nicholas J. Higham},
    title = {Functions of Matrices: Theory and Computation},
    publisher = {SIAM},
    address = {Philadelphia, PA, USA},
    year = {2008},
    pages = {xx+425},
}

@article{higham1986,
    author = {Higham, Nicholas J.},
    title = {Computing the Polar Decomposition-with Applications},
    journal = {SIAM Journal on Scientific and Statistical Computing},
    volume = {7},
    number = {4},
    pages = {1160--1174},
    year = {1986},
    doi = {10.1137/0907079},
}

@article{higham1990,
    author = {Higham, Nicholas J. and Schreiber, Robert S.},
    title = {Fast Polar Decomposition of an Arbitrary Matrix},
    journal = {SIAM Journal on Scientific and Statistical Computing},
    volume = {11},
    number = {4},
    pages = {648--655},
    year = {1990},
    doi = {10.1137/0911038},
}

@article{nakatsukasa2013,
    title = {Stable and Efficient Spectral Divide and Conquer Algorithms for the Symmetric Eigenvalue Decomposition and the SVD},
    volume = {35},
    doi = {10.1137/120876605},
    number = {3},
    journal = {SIAM Journal on Scientific Computing},
    author = {Nakatsukasa, Yuji and Higham, Nicholas J.},
    month = {1},
    year = {2013},
    pages = {A1325--A1349},
}

@article{nakatsukasa2016,
    title = {Computing Fundamental Matrix Decompositions Accurately via the Matrix Sign Function in Two Iterations: The Power of {Zolotarev}'s Functions},
    volume = {58},
    doi = {10.1137/140990334},
    number = {3},
    journal = {SIAM Review},
    author = {Nakatsukasa, Yuji and Freund, Roland W.},
    month = {1},
    year = {2016},
    pages = {461--493},
}

@phdthesis{huang2024phd,
  author = {Hua Huang}, 
  title = {New Parallel Algorithms for Large-Scale Matrix Computations},
  school = {Georgia Institute of Technology},
  year = {2024},
  month = {July},
}

@misc{nvidia2025nemotron-h,
    title = {{Nemotron-H}: A Family of Accurate and Efficient Hybrid Mamba-Transformer Models}, 
    author={NVIDIA and Aaron Blakeman and Aarti Basant and Abhinav Khattar and Adithya Renduchintala and Akhiad Bercovich and Aleksander Ficek and Alexis Bjorlin and Ali Taghibakhshi and Amala Sanjay Deshmukh and Ameya Sunil Mahabaleshwarkar and Andrew Tao and Anna Shors and Ashwath Aithal and Ashwin Poojary and Ayush Dattagupta and Balaram Buddharaju and Bobby Chen and Boris Ginsburg and Boxin Wang and Brandon Norick and Brian Butterfield and Bryan Catanzaro and Carlo del Mundo and Chengyu Dong and Christine Harvey and Christopher Parisien and Dan Su and Daniel Korzekwa and Danny Yin and Daria Gitman and David Mosallanezhad and Deepak Narayanan and Denys Fridman and Dima Rekesh and Ding Ma and Dmytro Pykhtar and Dong Ahn and Duncan Riach and Dusan Stosic and Eileen Long and Elad Segal and Ellie Evans and Eric Chung and Erick Galinkin and Evelina Bakhturina and Ewa Dobrowolska and Fei Jia and Fuxiao Liu and Gargi Prasad and Gerald Shen and Guilin Liu and Guo Chen and Haifeng Qian and Helen Ngo and Hongbin Liu and Hui Li and Igor Gitman and Ilia Karmanov and Ivan Moshkov and Izik Golan and Jan Kautz and Jane Polak Scowcroft and Jared Casper and Jarno Seppanen and Jason Lu and Jason Sewall and Jiaqi Zeng and Jiaxuan You and Jimmy Zhang and Jing Zhang and Jining Huang and Jinze Xue and Jocelyn Huang and Joey Conway and John Kamalu and Jon Barker and Jonathan Cohen and Joseph Jennings and Jupinder Parmar and Karan Sapra and Kari Briski and Kateryna Chumachenko and Katherine Luna and Keshav Santhanam and Kezhi Kong and Kirthi Sivamani and Krzysztof Pawelec and Kumar Anik and Kunlun Li and Lawrence McAfee and Leon Derczynski and Lindsey Pavao and Luis Vega and Lukas Voegtle and Maciej Bala and Maer Rodrigues de Melo and Makesh Narsimhan Sreedhar and Marcin Chochowski and Markus Kliegl and Marta Stepniewska-Dziubinska and Matthieu Le and Matvei Novikov and Mehrzad Samadi and Michael Andersch and Michael Evans and Miguel Martinez and Mike Chrzanowski and Mike Ranzinger and Mikolaj Blaz and Misha Smelyanskiy and Mohamed Fawzy and Mohammad Shoeybi and Mostofa Patwary and Nayeon Lee and Nima Tajbakhsh and Ning Xu and Oleg Rybakov and Oleksii Kuchaiev and Olivier Delalleau and Osvald Nitski and Parth Chadha and Pasha Shamis and Paulius Micikevicius and Pavlo Molchanov and Peter Dykas and Philipp Fischer and Pierre-Yves Aquilanti and Piotr Bialecki and Prasoon Varshney and Pritam Gundecha and Przemek Tredak and Rabeeh Karimi and Rahul Kandu and Ran El-Yaniv and Raviraj Joshi and Roger Waleffe and Ruoxi Zhang and Sabrina Kavanaugh and Sahil Jain and Samuel Kriman and Sangkug Lym and Sanjeev Satheesh and Saurav Muralidharan and Sean Narenthiran and Selvaraj Anandaraj and Seonmyeong Bak and Sergey Kashirsky and Seungju Han and Shantanu Acharya and Shaona Ghosh and Sharath Turuvekere Sreenivas and Sharon Clay and Shelby Thomas and Shrimai Prabhumoye and Shubham Pachori and Shubham Toshniwal and Shyamala Prayaga and Siddhartha Jain and Sirshak Das and Slawek Kierat and Somshubra Majumdar and Song Han and Soumye Singhal and Sriharsha Niverty and Stefania Alborghetti and Suseella Panguluri and Swetha Bhendigeri and Syeda Nahida Akter and Szymon Migacz and Tal Shiri and Terry Kong and Timo Roman and Tomer Ronen and Trisha Saar and Tugrul Konuk and Tuomas Rintamaki and Tyler Poon and Ushnish De and Vahid Noroozi and Varun Singh and Vijay Korthikanti and Vitaly Kurin and Wasi Uddin Ahmad and Wei Du and Wei Ping and Wenliang Dai and Wonmin Byeon and Xiaowei Ren and Yao Xu and Yejin Choi and Yian Zhang and Ying Lin and Yoshi Suhara and Zhiding Yu and Zhiqi Li and Zhiyu Li and Zhongbo Zhu and Zhuolin Yang and Zijia Chen},
    year={2025},
    eprint={2504.03624},
    archivePrefix={arXiv},
    primaryClass={cs.CL},
    url={https://arxiv.org/abs/2504.03624}, 
}

@misc{hagele2024,
    title={Scaling Laws and Compute-Optimal Training Beyond Fixed Training Durations}, 
    author={Alexander Hägele and Elie Bakouch and Atli Kosson and Loubna Ben Allal and Leandro Von Werra and Martin Jaggi},
    year={2024},
    eprint={2405.18392},
    archivePrefix={arXiv},
    primaryClass={cs.LG},
    url={https://arxiv.org/abs/2405.18392}, 
}

\end{document}